\pdfoutput=1
\documentclass[conference]{IEEEtran}
\IEEEoverridecommandlockouts
\usepackage{cite}
\usepackage{amsmath,amssymb,amsfonts}
\usepackage{algorithmicx}
\usepackage{algorithm} 
\usepackage{algpseudocode}

\usepackage{graphicx}
\usepackage{textcomp}
\usepackage{xcolor}
\def\BibTeX{{\rm B\kern-.05em{\sc i\kern-.025em b}\kern-.08em
    T\kern-.1667em\lower.7ex\hbox{E}\kern-.125emX}}
\begin{document}

\title{An Ontological Knowledge Representation \\ for Smart Agriculture }

\author{\IEEEauthorblockN{Bikram Pratim Bhuyan\IEEEauthorrefmark{1}, Ravi Tomar\IEEEauthorrefmark{2}, Maanak Gupta\IEEEauthorrefmark{3}, and Amar Ramdane-Cherif\IEEEauthorrefmark{4}}
\IEEEauthorblockA{\IEEEauthorrefmark{1}\IEEEauthorrefmark{2}Dept. of Informatics, School of Computer Science, \\ University of Petroleum and Energy Studies, Dehradun, India \\
\IEEEauthorrefmark{3}{Dept. of Computer Science},
{Tennessee Technological University},
Cookeville, Tennessee 38505, USA \\\IEEEauthorrefmark{4}LISV Laboratory, University of Versailles
Versailles, France\\}
\IEEEauthorrefmark{1}bikram23bhuyan@gmail.com,
\IEEEauthorrefmark{2}ravitomar7@gmail.com, 
\IEEEauthorrefmark{3}mgupta@tntech.edu
\IEEEauthorrefmark{4}amar.ramdane-cherif@uvsq.fr}

\maketitle

\begin{abstract}
In order to provide the agricultural industry with the infrastructure it needs to take advantage of advanced technology, such as big data, the cloud, and the internet of things (IoT); smart farming is a management concept that focuses on providing the infrastructure necessary to track, monitor, automate, and analyse operations. To represent the knowledge extracted from the primary data collected is of utmost importance. An agricultural ontology framework for smart agriculture systems is presented in this study. The knowledge graph is represented as a lattice to capture and perform reasoning on spatio-temporal agricultural data. 
\end{abstract}


\section{Introduction}
People all across the globe depend on agriculture for their daily sustenance.  Crop yields are decreasing, which drives up food costs~\cite{b2}. Additionally, as the ordinary person becomes wealthier, their eating habits are changing and they are consuming more food, including more meat. Land, water, energy, and other resources are becoming more scarce, making it difficult to meet the world's growing need for food. Water contamination, poor soil fertility, fertiliser misuse, climate change, or illnesses may push a country's population below the poverty line. Internet of things (IOT) in conjunction with wireless sensor networks (WSN) is seen as the possible answer for making effective interventions in agriculture. Agriculture is being revolutionised by the usage of IoT, a combination of developing technologies that gives farmers the tools they need to meet the immense problems of the twenty-first century. Precision agriculture (PA), commonly referred to as site-specific farming, is a kind of agricultural management that relies heavily on information.

Several works \cite{b26,b27,b28,b29,b30} have developed to emphasize the need for smart agriculture systems along with associated cybersecutity  concerns. Throughout human history, significant advancements have been developed to increase agricultural productivity while using less resources and work hours. Despite this, the high population rate has always prevented supply and demand from matching. According to projections, the global population will reach 9.8 billion people in 2050, up by almost 25\% from today's level~\cite{b3}. Crop production is essential not just for food, but also for industry; in fact, crops like cotton, rubber, and gum are key to the economy of many countries. In addition, the food-crops-based bioenergy sector has lately begun to grow in size. In-depth research revealed that each agricultural field has unique features that may be assessed independently in terms of both quality and quantity. Soil type, nutrient present, irrigation flow, insect resistance, and other critical elements establish a crop's appropriateness and capacity. Even though the same crop is being planted over the whole farm, differences in features might occur within a single field in most cases, necessitating site-specific analysis in order to provide the best possible yield~\cite{b4}.

Various methodologies and architectures are being proposed by scientists and engineers all over the world based on which a wide range of equipment is being suggested to monitor and retrieve information on crop status at various stages, taking a diversity of crop and field kinds into consideration. The market need is driving several prominent manufacturers to provide sensors, unmanned aerial vehicles (UAV) and other heavy gear to transport data collected by the robots~\cite{b5}. In this paper, we formally formulate the representation of the data collected in a lattice structure and build a mining algorithm for rule generation.

\section{Literature Survey}
Multitude of work over the use of IOT in agriculture can be found in literature.  The chapter~\cite{b6} gives an overview of IoT applications, including where they are now and where they are going in the future. Specific IoT-based PA applications, including as precision irrigation, fertilisation, agricultural disease management, and pest control, are discussed in detail in the various parts of the book. Articles in significant journals were reviewed in~\cite{b7}, current research trends were analysed, and common IoT sensors and devices were investigated. Agriculture APPs, advantages and problems in IoT-based agricultural production were also investigated. Accordingly, the paper stated that smart agriculture's future developments include building a generic platform for all kinds of crops and animals, QoS (Quality of Service), and using explainable artificial intelligence to monitor crop growth and disease prevention.

Experiment was conducted on agricultural equipment vibration and tilt using five different machine learning algorithms in~\cite{b8} (K-Nearest Neighbor; Support Vector Machine; Decision Tree; Random Forest; and Gradient Boosting). The results from Random Forest were the most precise. Over-fitting was seen in the Gradient boosting and Random Forest algorithms; yet, they both generated the greatest testing accuracies.

When it comes to precision agriculture, there are a number of different ontologies to consider. AGROVOC~\cite{b9} is the most often used ontology. Because of its inconsistently assigned connections and too wide definitions, AGROVOC is simply a decent vocabulary system to begin creating an agricultural ontology, not an ontology for agriculture itself. AgriOnt is a framework for smart agriculture presented by the authors in~\cite{b1}. Agri-based ontology is divided into four parts: geographical ontology, the business domain, the IoT domain, and an agricultural domain. A framework was presented, however it is constrained by context awareness, scalability, and service-oriented design. This is a challenge.

The researchers in~\cite{b10} came up with SAAONT, which is targeted towards Saudi Arabian farmers. Earlier ontologies couldn't provide farmers with the information they needed. SAAONT's structure implementation uses many steps of the ontology creation process, including ontology selection, re-use, modelling, and interaction design. Because the Arabic language is not widely recognised by farmers throughout the globe, this ontology is restricted to the Arabic area alone. Authors solely targeted Saudi Arabia and built an ontology for the Arabic region.

Site-specific characteristics were linked to crop-specific trait ontologies in~\cite{b11} to sensor measurement data. Main characteristic of the technique is data integration enabling syntactic and semantic compatibility. Use of APIs and web services in the agricultural area for syntactic interoperability of gadget information is an important aspect of the investigation. An ontology model was used by the authors in~\cite{b12} to create a recommendation system for describing insect outbreaks in crops and the acceptable methods of treating and facilitating them. Researchers in this study also examine pests, crops, and the disease triangle's treatment ontologies (PCT-O). As an example, the ontology of knowledge about longan fruit farming and production was constructed by the authors in~\cite{b13} and used using a semantic search approach in order to improve fruit quality and cultivate quality longan fruit fit for the worldwide market.

Crop cultivation standard (CCS) and task ontology and domain ontology are used by the authors~\cite{b14} to express a technique. Static information pertaining to domain ontologies, such seeds, soils, and agricultural machinery, makes up the bulk of the data. For example, soil and seed selection, irrigation and fertilisation are all part of the task ontology. Climate change ontologies were studied in~\cite{b15}.

SAGRO-Lite, a lightweight ontology built by the authors in~\cite{b16} for unique agricultural features in poor nations, has been used by the authors to construct a semantically enhanced agent-based model dubbed Agent Based Semantic Model for Smart Agriculture, ABSMSA. For semantic sensing and event detection and processing, the system makes use of two additional ontologies: IoT-Lite and the Complex Event Service Ontology (CESO). A rapid uptake of IoT in farming is not feasible in low- and middle-income countries due to literacy issues, apprehension about new technology, a lack of available land, and the high cost of IoT agricultural solutions. Lightweight Internet of Things (IoT) tailored to developing country agricultural practises, such as in India, may improve farmers' productivity and yield.

Using Boosted Continuous Non-spatial whole-attribute Extraction, the authors in~\cite{b17} proposed an ontology enabled IoT for distinguishing healthy separation of sekai-ichi apples. The Post-Harvest procedure wastes a significant amount of Sekai-ichi apple due to its propensity to spread disease. It's particularly risky to lose the sekai-ichi apple count since it doesn't spoil quickly after harvest. It was suggested that a post-harvest hierarchical model be used to specify post-harvest losses prevention and deficiencies to precisely and quickly identify waste in order to maintain a healthy separation of agriculture from the surrounding environment. There are three levels of processing schemes used in these recommendations, with the separation cognitive operation taking the three schemes from lowest to highest.

A few scholars worked on ontologies, while others constructed recommendation systems, provided case studies, and created models and frameworks in the previously stated linked work. The major issues of this studies are service-oriented, scalable, and aware of the context. The depiction of temporal and spatial data remains a challenge as well. Taking these pitfalls in the previous works into consideration, we tend to present a spatio-temporal ontological structure for knowlwdge representation. The spatio-temporal data collected from the IoT devices are processed to form a lattice structure which is farther used to generate rules based on the lattice properties.  Formal definition for the knowledge representation of such spatio-temporal agricultural data is carried out in the next section.

\section{Knowledge Representation and Reasoning}

For automated reasoning, knowledge representation~\cite{b18} in artificial intelligent (AI) refers to the process of creating a clear symbolic notation that expresses an intelligent agent's beliefs, intents, and values. From a strictly computational perspective, the primary goals to be fulfilled are scope, expressivity, accuracy, support for efficient inference, learnability, resilience and simplicity of development. A wide range of computing problems, including text interpretation and cognitive robots, have benefited from knowledge-based strategies~\cite{b19}. While suitable knowledge representations are important for software design and performance, our choice of knowledge representation systems also exposes our implicit ideas about the nature of machine and human intelligence. If rule-based representations and semantic networks have different ideas on how knowledge is structured in human brains, then it's possible that the two camps will disagree~\cite{b20}. 

To represent knowledge, a variety of generic designs have been utilised, including first-order logic, formal logics, semantic networks, and frame-based systems~\cite{b21}. In constructing representations for many domains, the difficulty of representing temporal information is an example of the types of difficulties that might occur~\cite{b31}. Knowledge management representation languages should be simple to use, even for non-specialists, and promote human communication, which is frequently facilitated by the ability to express still-unrefined concepts, reduce complexity, omit specifics, and host several points of view all at once. All of these qualities may be found in graphic representation languages, which are commonly employed to enable information flow.

We now work with some basic definitions for the formal knowledge representation.

\section{Basic Definitions}
As data collected from the devices are spatio-temporal in nature, we have we work with data cubes or in other sense `panel data'. 
\subsection{Panel Data}
When talking about longitudinal data, it's common to use the term `panel data'~\cite{b22}, which refers to a collection of observations from several time periods. Panel data, like time series data, is made up of observations that are gathered on a regular basis and organised chronologically. Observations on a large group of people are found in panel data, much as in cross-sectional data. It's a special area of time series modelling to model these panel data series since they have a distinctive structure~\cite{b24}. There are two main types of panel data methods~\cite{b23}: a) This model assumes that all people have the same set of model parameters, which is known as the homogeneous assumption. b) Heterogeneous models, on the other hand, allow for individual differences in any or all of the model parameters. Heterogeneous panel data models include models with fixed effects and random effects. 
\subsection{Formal Interpretation}
Each location of the devices to collect data is represented as a set $L = \{L_1, L_2, \ldots, L_n\}$ where `n' is the total number of distinct location called as `agro-location'. The constantly collected information or data which is denoted by $J = \{J_1, J_2, \ldots, J_k\}$ where `k' is the total distinct properties or dimensions captured by the smart device. As data is collected across various timestamps, it is denoted by $T = \{T_1, T_2, \ldots, T_t\}$ where `t' is the number of instances the data is collected. Table~\ref{tab1} gives a reference to the symbols used in this article.

\begin{table}[htbp]
\caption{Notations used in the article}
\begin{center}
\begin{tabular}{|c|c|}
\hline
\textbf{Hieroglyph}&{\textbf{Interpretation}} \\
\hline
$L$ & Set of distinct agro-locations  \\
\hline
$J$ & Set of distinct dimensions  \\
\hline
$T$ & Set of distinct timestamps  \\
\hline
$P$ & Projection function  \\
\hline
$l^{\oslash}$ & Friendly function on a set $l \subseteq L$  \\
\hline
$j^{\oslash}$ & Friendly function on a set $j \subseteq J$  \\
\hline
$t^{\oslash}$ & Friendly function on a set $t \subseteq T$  \\
\hline
$\partial (L, J, T)$ & Spatio-temporal lattice structure  \\
\hline
$(L, J, T, P)$ & Spatio-temporal structure  \\
\hline
$(l, j, t)$ & Agro-triple  \\
\hline
$\vee$ & Lattice suprema \\
\hline
$\wedge$ & Lattice infima  \\
\hline
$\phi$ & Support count  \\
\hline
$\mu$ & Confidence  \\
\hline
$\Delta$ & Set of associative rules  \\
\hline
\end{tabular}
\label{tab1}
\end{center}
\end{table}

We are now set to define a projection function to map locations to their respective property in the timestamp.

\textbf{Definition 1.} A projection function $P$ is defined from the agro location to the properties with a timestamp as:
\begin{equation}
P : L \mapsto J \times T
\label{eq}
\end{equation}
The projection function operates amongst the two sets as a binary operator to capture the presence of a certain dimension or property at that particular time for the agro-location (denoted by `c').

A slice of the data-cube is shown in Table~\ref{tab2}, where for instance $L_n$ possess the dimension set $\{J_1, J_2, J_k\}$ for a distinct time-stamp. Similar slice is shown in Table~\ref{tab3} where for example, the presence of the dimension $J_2$ is seen across the timestamp set $\{T_1, T_t\}$ for the same agro-location.

\begin{table}[htbp]
\caption{Projection from agro-location to dimensions}
\begin{center}
\begin{tabular}{|c|c|c|c|c|}
\hline
\textbf{Agro-Location}&\textbf{$J_1$} &{\textbf{$J_2$}} &{\textbf{$\ldots$}}&{\textbf{$J_k$}} \\
\hline
$L_1$ & c &c & &  \\
\hline
$L_2$ &  &c & &c  \\
\hline
$L_3$ & c & &c &  \\
\hline
$L_4$ &  & &c &c  \\
\hline
$\ldots$ & c & & &  \\
\hline
$L_n$ & c &c & &c  \\
\hline

\end{tabular}
\label{tab2}
\end{center}
\end{table}

\begin{table}[htbp]
\caption{Projection from timestamp to dimensions}
\begin{center}
\begin{tabular}{|c|c|c|c|c|}
\hline
\textbf{Timestamp}&\textbf{$J_1$} &{\textbf{$J_2$}} &{\textbf{$\ldots$}}&{\textbf{$J_k$}} \\
\hline
$T_1$ &  &c & &c  \\
\hline
$T_2$ &  & & &c  \\
\hline
$T_3$ & c & & &  \\
\hline
$T_4$ &  & &c &  \\
\hline
$\ldots$ & c & &c &  \\
\hline
$T_t$ &  &c & &c  \\
\hline

\end{tabular}
\label{tab3}
\end{center}
\end{table}

\textbf{Definition 2.} A friendly function $l^{\oslash}$ on a set $l \subseteq L$ is defined as:
\begin{equation}
l^{\oslash} = (j, t) \in \{(J \times T) | (l, m, t) \in P, \forall l \in L\}
\label{eq2}
\end{equation}
where $j \subseteq J$ and $t \subseteq T$.

Thus, this friendly operator on a subset of the agro-location captures the tuples in both space and time where the common dimensions with the respective timestamp for that set of location is collected and stored. Similar friendly functions can be created for the dimensions and timestamp as follows.

\textbf{Definition 3.} A friendly function $j^{\oslash}$ on a set $j \subseteq J$ is defined as:
\begin{equation}
j^{\oslash} = (l, t) \in \{(L \times T) | (l, m, t) \in P, \forall j \in J\}
\label{eq3}
\end{equation}
where $l \subseteq L$ and $t \subseteq T$.

This function defines the tuples in terms of the location and timestamp which are common to the set of dimensions or properties.

\textbf{Definition 4.} A friendly function $t^{\oslash}$ on a set $t \subseteq T$ is defined as:
\begin{equation}
t^{\oslash} = (l, j) \in \{(L \times J) | (l, m, t) \in P, \forall t \in T\}
\label{eq4}
\end{equation}
where $l \subseteq L$ and $j \subseteq J$.

This function captures the tuples which are common in the time set in regards to the set of agro-location and dimension. Thus equations \ref{eq2}-\ref{eq4} manages to combine the common tuples in all possible ways.

A set is said to fulfil a closure property~\cite{b25} if one operation or a series of operations closes the data in the set. It's not uncommon for axioms to add closure properties as properties of their own. Adding closure to a structure as an axiom is totally pointless in modern set-theoretic definitions because operations are usually defined as maps between sets. However, in practise, operations are often defined first on a superset of the set in question, and a closure proof is needed to establish that the functionality implemented to combinations from that set only produces members of that set An even integer collection is closed under addition, whereas an odd integer collection is not.

\textbf{Lemma 1.} The familiarity functions are closed under the same function.

The above Lemma states that:
\begin{equation}
    l^{\oslash \oslash} = l
\end{equation}

\begin{equation}
    j^{\oslash \oslash} = j
\end{equation}

\begin{equation}
    t^{\oslash \oslash} = t
\end{equation}

\textbf{Definition 5.} A spatio-temporal structure is defined as the tuple $(L, J, T, P)$. Where `L' is the set of agro-location. `J' is the set of dimensions or properties collected from those locations, `T' is the set of temporal timestamps over the period which the data is collected and `P' is the mapping from the set of agro-location to the collection of property with timestamp. 

\textbf{Definition 6.} An agro-triple is defined as $(l, j, t)$. Where $l \subseteq L$, $j \subseteq J$ and $t \subseteq T$.

Due to the closure property seen in equations~5, 6 and 7 and the friendly functions in equations~2, 3 and 4; an agro-triple can also be defined as $(j^{\oslash}, j^{\oslash \oslash}, t)$

After the formal definitions, we are now at a state to formally design some algorithms for the knowledge graph creation.

\section{Knowledge Representation for Smart Agriculture}
In order to form a knowledge representation in terms of a lattice for smart agriculture, we start off with an agglomeration algorithm 1.

\begin{algorithm}
\label{Conceptcreate}
 \caption{Agglomeration of agri-location with dimension and timestamp}
 Combine location, dimension and timestamp to form a triple $(l, j, t)$ where $l \subseteq L$, $j \subseteq J$ and $t \subseteq T$.
 
 Input: Set of agri-location: $l \subseteq L$;  Set of dimensions: $j \subseteq J$ and Set of timestamps: $t \subseteq T$

Output: Triples set (TS)
 \begin{algorithmic}[1]
 
\State{Create\textendash Triples}()
\State Initialize $l_0$, $j_0$ and $t_0$ to $\phi$.
\For{each timestamp $t_i$ $\in$ $T$}
\For{each agri-location $l_i$ $\in$ $L$}
\State Compute ${l_i}^{\oslash}$ and ${j_i}^{\oslash\oslash}$
\State \textbf{if} ${l_i}^{\oslash}$ == ${j_i}^{\oslash\oslash}$
    \State $L_i \leftarrow {l_i}^{\oslash\oslash}$
    \State $J_i \leftarrow {j_i}^{\oslash}$
    \State Form tuple $(L_i, J_i)$ for each $t_i$
    \State Remove duplicates if any.
\EndFor
\State Compute $t_i^{\oslash}$ and $(L_i, J_i)^{\oslash\oslash}$
\State \textbf{if} ${t_i}^{\oslash}$ == $(L_i, J_i)^{\oslash\oslash}$
    \State $(L, J) \leftarrow (L_i, J_i)^{\oslash\oslash}$
    \State $T \leftarrow {t_i}^{\oslash}$
    \State Form triple $(L, J, T)$
    \State Remove duplicates if any.
\EndFor
 \end{algorithmic} 
 \end{algorithm}

Algorithm~1 deals with creating triples $(L, J, T)$ forming a closed set due to the properties seen in Lemma~1. The triples thus collected are kept in a set for further analysis and representation. The lattice representation of the collection is formulated using Algorithm~2. 
 \begin{algorithm}[h!]
 \caption{Lattice Structure}
 Lattice construction from the triples collected.  
 
 Input: Spatio-temporal Structure 

Output: Spatio-temporal Lattice 
 \begin{algorithmic}[1]
 
\Procedure{Construct\textemdash Agro-triple}{}
\For{each dimension $j \in$ $J$ at time $t \in$ $T$ }
\State Find $j_{t}^{\oslash}$ and $j_{t}^{\oslash \oslash}$
\State Form agro-triple $(j_{t}^{\oslash}, j_{t}^{\oslash \oslash}, T_t)$
\EndFor
 \EndProcedure
\Procedure{Construct\textemdash Lattice Structure}{}
\For{each agro-triple (TS)}
  \State \textbf{if} (($j_{x_{t}}^{\oslash}$ $>$ $j_{y_{t}}^{\oslash}$) $\&\& $ ($j_{x_{t}}^{\oslash \oslash}$ $<$ $j_{y_{t}}^{\oslash \oslash}$) $\&\& $ ($t_x$ $<$ $t_y$)) 
  \State Position ($j_{x_{t}}^{\oslash}$, $j_{x_{t}}^{\oslash \oslash}$, $t_x$) as the super triple of ($j_{y_{t}}^{\oslash}$, $j_{y_{t}}^{\oslash \oslash}$, $t_y$)
  \State Construct an edge from ($j_{x_{t}}^{\oslash}$, $j_{x_{t}}^{\oslash \oslash}$, $t_x$) which is hierarchically above to ($j_{y_{t}}^{\oslash}$, $j_{y_{t}}^{\oslash \oslash}$, $t_y$).
  \EndFor
 \EndProcedure
 \end{algorithmic} 
 \label{S_algorithm}
 \end{algorithm}

Thus, algorithm~2 creates a meet lattice from the triples created from algorithm~1. We now provide some definitions relating to the same.

\textbf{Definition 7.} An agro-triple $(l, j, t)$ is a triple which is created from a spatio-temporal structure $(L, J, T, P)$ and collectively forms a lattice structure `$\partial (L, J, T)$'.

Order theory and abstract algebra both study lattices as abstract structures. Partly ordered set where each pair of elements has a unique supremum (also known as a least upper limit or join) and unique infimum (also called a greatest lower bound or meet).

\textbf{Definition 8.} An agro-triple $(l_x, j_x, t_x)$ is said to be a super triple of $(l_x, j_x, t_x)$ iff $l_y \subseteq l_x$, $j_x \subseteq j_y$ and $t_x \subseteq t_y$. This set of agro-triples in such hierarchy finally forms the lattice structure `$\partial (L, J, T)$'.

\textbf{Definition 9.} The lattice structure $\partial (L, J, T)$ fulfills the partial ordered property of a lattice bearing a lattice suprema defined as:

\begin{equation}
    \vee_{t \in T} < (l, j, t) > = < \cap_{t \in T} l^{\oslash \oslash}, \cup_{t \in T} j, \cup_{t \in T} t  >
\end{equation}

\textbf{Definition 10.} The lattice structure $\partial (L, J, T)$ fulfills the partial ordered property of a lattice bearing a lattice infima defined as:

\begin{equation}
    \wedge_{t \in T} < (l, j, t) > = < \cup_{t \in T} l, \cap_{t \in T} j^{\oslash \oslash}, \cap_{t \in T} t  >
\end{equation}

\textbf{Definition 11.} The partial ordered property along with the property that each and every agro-triple has both supremum (join) and infimum (meet), guarantees $\partial (L, J, T)$ to be a complete lattice. 

After the formal definitions of the lattice structure formed from agro-triples, we now concentrate on the rules that could be generated from the same. 

Associative rule learning is a machine learning technique based on rules for uncovering new relationships between variables in huge datasets. Its goal is to find interesting rules in databases and identify the ones that are powerful. To uncover common patterns, correlations, connections, or causal structures in different types of databases such as relational databases, transactional databases, and other types of data repositories, a method known as association rule mining is used. If we have a series of transactions, association rule mining attempts to uncover those rules that let us know when something will happen by looking at what else has happened in the transaction.

\textbf{Definition 12.} The agro-triples generates associative rules amongst the dimensions for a timestamp. Formally,

\begin{equation}
    \forall(l, j, t) \in \partial (L, J, T); \exists j_x \wedge j_y ; j_x \xrightarrow{t} j_y
\end{equation}
Where, $j_x$, $j_y$ $\subseteq j$ and $j_x \cup j_y =  j$

If-then statements in association rules offer this kind of information. Unlike logical if-then rules, association rules are generated from data and are, as a result, probabilistic in nature. There are two numbers in an association rule that describe the degree of ambiguity about the rule in addition to the antecedent (if) and the consequent (then). Association analysis uses groups of items (called itemsets) that are distinct as the antecedent and consequent (do not have any items in common). The first number is known as the rule's support. The basis for the rule's support is the total number of transactions, which includes all items in the preceding and following portions. There are occasions when the support is reported as a percentage of the database's total records. The second number is referred to as the rule's level of confidence. It is the ratio of the number of transactions that contain all the subsequent and antecedent (the support) items to the number of transactions that include all the antecedent.

\textbf{Definition 13.} The support $\phi$ of each rule $j_x \xrightarrow{t} j_y$ genereted from the agro-triples (l, j, t) of a spatio-temporal lattice structure $\partial (L, J, T)$ is defined as:
\begin{equation}
    \phi (j_x \xrightarrow{t} j_y) = \frac{|\{j_x \cup_t j_y\}^\oslash|}{|J|} = \frac{|l_t|}{|J|}
\end{equation}

Similarly, we define confidence as;

\textbf{Definition 14.} The confidence $\mu$ of each rule $j_x \xrightarrow{t} j_y$ genereted from the agro-triples (l, j, t) of a spatio-temporal lattice structure $\partial (L, J, T)$ is defined as:

\begin{equation}
    \mu (j_x \xrightarrow{t} j_y) = \frac{|\{j_x \cup_t j_y\}^\oslash|}{|j_x^\oslash|} = \frac{|l_t|}{|j_x^\oslash|}
\end{equation}

We now present a formal algorithm~3 to generate and store the associative rules.  

\begin{algorithm}[h!]
 \caption{Agro-triple rule generation}
Generate association rules from agro-triples from spatio-temporal lattice structure  
 
 Input: Set of Agro-triples 

Output: Set of associative rules
 \begin{algorithmic}[1]
 
\Procedure{Generate\textemdash associative-rules}{}
\State $\Delta= \phi$
\For{each agro-triple (l, j, t)}
  \State \textbf{if} (($j_{target}$ $\in$ $j$)  $\&\&$ ($t_{target}$ $\in$ $t$)) 
  \State Generate rules $\Delta_i :$ $j \xrightarrow{t}$ $j_{target}$ 
  \State $\Delta= \Delta \bigcup \Delta_i$
  \EndFor
  \State Compute $\mu$ and $\phi$ for each rule in $\Delta$
 \EndProcedure
 \end{algorithmic} 
 \label{S1_algorithm}
 \end{algorithm}

The representation of agro-data as a data cube can be viewed as shown in Table~\ref{tab4} or as Table~\ref{tab5}. A few statements can be observed.

\begin{table}[htbp]
\caption{Location-Dimension projection with timestamp}
\begin{center}
\begin{tabular}{|c|c|c|c|c|c|c|c|c|c|}

 \multicolumn{1}{c}{} & \multicolumn{4}{c}{$T_1$} & \multicolumn{1}{c}{\ldots} & \multicolumn{4}{c}{$T_t$}
\\
\hline
  & $J_1$ & $J_2$ & \ldots & $J_k$ & \ldots & $J_1$ & $J_2$ & \ldots & $J_k$ 
 \\ 
\hline
$L_1$ & c & c & c  &   & c  &   & c &   &    
\\ 
\hline
$L_2$ &  & c & c  &   &   & c  & c & c & c  
\\ 
\hline
\ldots &  &  & c & c  & c  &   &  &  & c  
\\
\hline
$L_n$ & c & c  &  & c  & c  &   &  & c & c  
\\
\hline

\end{tabular}
\label{tab4}
\end{center}
\end{table}

\begin{table}[htbp]
\caption{Temporal-Location projection with dimension}
\begin{center}
\begin{tabular}{|c|c|c|c|c|c|c|c|c|c|}

 \multicolumn{1}{c}{} & \multicolumn{4}{c}{$J_1$} & \multicolumn{1}{c}{\ldots} & \multicolumn{4}{c}{$J_k$}
\\
\hline
  & $T_1$ & $T_2$ & \ldots & $T_t$ & \ldots & $T_1$ & $T_2$ & \ldots & $T_t$ 
 \\ 
\hline
$L_1$ & c & c &   &   &   &   & c &   &    
\\ 
\hline
$L_2$ & c & c &   &   &   &   & c & c &   
\\ 
\hline
\ldots & c & c & c &   &   &   & c & c &   
\\
\hline
$L_n$ & c &   & c &   &   &   & c & c & c  
\\
\hline

\end{tabular}
\label{tab5}
\end{center}
\end{table}

\textbf{Statement 1.} Homogeneous agro-triples are generated from both of the data cubes shown in Table~\ref{tab4} and Table~\ref{tab5} with the identical count. 

\textbf{Statement 2.} Isomorphic spatio-temporal lattice structure is generated from both of the projections.

Hence, regardless of the raw agro-data projection as a data cube, we can first generate the agro-triples and then create the spatio-temporal lattice structure which will finally lead to a set associative rules for reasoning. Thus we have constructed a knowledge graph in terms of a lattice for knowledge representation and reasoning.

\section{Experimental Results and Discussion}
We test our algorithms with the help of toy datasets created for the purpose as shown in Table~\ref{tab6} - ~\ref{tab9}. We have taken into account ten agro-locations $\{L_1, L_2, \ldots, L_{10}\}$. Six dimensions or properties are taken $\{J_1, J_2, \ldots, J_{6}\}$ over a time span of four timestamps $\{T_1, T_2, \ldots, T_{4}\}$.   

\begin{table}[htbp]
\caption{Toy data set Projection from agro-location to dimensions($T_1$)}
\begin{center}
\begin{tabular}{|c|c|c|c|c|c|c|c}
\hline
\textbf{Agro-Location}&\textbf{$J_1$} &{\textbf{$J_2$}} &{\textbf{$J_3$}}& $J_4$ & $J_5$& {\textbf{$J_6$}} \\
\hline
$L_1$ & c &c & & c& c & \\
\hline
$L_2$ &  &c & c& c& c & c \\
\hline
$L_3$ &  & & c& & c & c \\
\hline
$L_4$ & c &c & & & c &  \\
\hline
$L_5$ & c &c & c& c& c &  \\
\hline
$L_6$ &  &c & & c&  & c\\
\hline
$L_7$ & c &c & c& c&  &  \\
\hline
$L_8$ &  & & & c& c & c \\
\hline
$L_9$ & c &c & c & &  &  \\
\hline
$L_{10}$ &  & &c & c& c &  \\
\hline

\end{tabular}
\label{tab6}
\end{center}
\end{table}

\begin{table}[htbp]
\caption{Toy data set Projection from agro-location to dimensions($T_2$)}
\begin{center}
\begin{tabular}{|c|c|c|c|c|c|c|c}
\hline
\textbf{Agro-Location}&\textbf{$J_1$} &{\textbf{$J_2$}} &{\textbf{$J_3$}}& $J_4$ & $J_5$& {\textbf{$J_6$}} \\
\hline
$L_1$ &  &c &c & c& c & \\
\hline
$L_2$ & c &c & c& & c & c \\
\hline
$L_3$ &  & & &c & c & c \\
\hline
$L_4$ & c &c & c& &  &  \\
\hline
$L_5$ & c &c & & c& c & c \\
\hline
$L_6$ & c & & & c& c & c\\
\hline
$L_7$ &  &c & c& c& c & c \\
\hline
$L_8$ & c & c& & c& c& c \\
\hline
$L_9$ & c &c & c &c &  &  \\
\hline
$L_{10}$ & c & &c & c& c &  \\
\hline

\end{tabular}
\label{tab7}
\end{center}
\end{table}

\begin{table}[htbp]
\caption{Toy data set Projection from agro-location to dimensions($T_3$)}
\begin{center}
\begin{tabular}{|c|c|c|c|c|c|c|c}
\hline
\textbf{Agro-Location}&\textbf{$J_1$} &{\textbf{$J_2$}} &{\textbf{$J_3$}}& $J_4$ & $J_5$& {\textbf{$J_6$}} \\
\hline
$L_1$ &  &c & & c& c & c\\
\hline
$L_2$ & c &c & c& & c &  \\
\hline
$L_3$ &  &c & &c & c &  \\
\hline
$L_4$ & c &c & & &  & c \\
\hline
$L_5$ & c &c & & c&  & c \\
\hline
$L_6$ & c & & & c&  & c\\
\hline
$L_7$ &  &c & c& c& c &  \\
\hline
$L_8$ & c & c& & c& & c \\
\hline
$L_9$ & c &c & c &c &  & c \\
\hline
$L_{10}$ & c & & & c& c &  \\
\hline

\end{tabular}
\label{tab8}
\end{center}
\end{table}

\begin{table}[htbp]
\caption{Toy data set Projection from agro-location to dimensions($T_4$)}
\begin{center}
\begin{tabular}{|c|c|c|c|c|c|c|c}
\hline
\textbf{Agro-Location}&\textbf{$J_1$} &{\textbf{$J_2$}} &{\textbf{$J_3$}}& $J_4$ & $J_5$& {\textbf{$J_6$}} \\
\hline
$L_1$ &  &c & c& c& c & \\
\hline
$L_2$ & c &c & & & c &  \\
\hline
$L_3$ &  &c & &c &  & c \\
\hline
$L_4$ & c &c & & &  &  \\
\hline
$L_5$ & c &c & & c& c & c \\
\hline
$L_6$ & c &c & & c&  & \\
\hline
$L_7$ &  &c & c& & c &  \\
\hline
$L_8$ & c & & & c& & c \\
\hline
$L_9$ & c &c &  &c &  & c \\
\hline
$L_{10}$ & c & c& & c& c &  \\
\hline

\end{tabular}
\label{tab9}
\end{center}
\end{table}

\begin{figure}[!t]
\centerline{\includegraphics[scale=0.03]{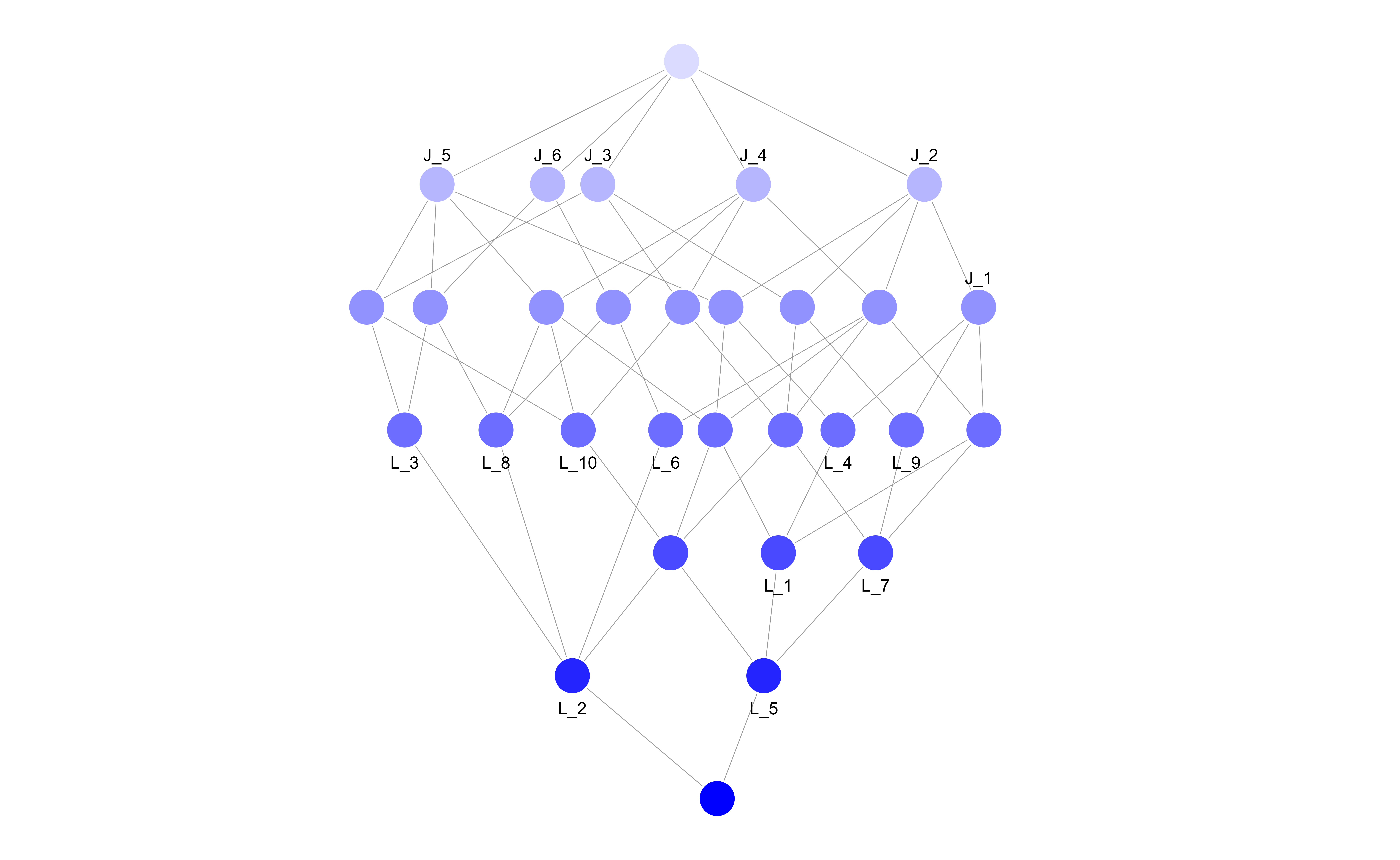}}
\caption{Spatio-Temporal Lattice Structure}
\label{fig}
\end{figure}

\begin{table}[htbp]
\caption{Agro-Triples generated}
\begin{center}
\begin{tabular}{|c|c|c|c|}
\hline
\textbf{Agro-Triple ID}&\textbf{Agro-Location}  &{\textbf{Dimension}} &{\textbf{Timestamp}} \\
\hline
$TS_1$ & $L_1$ & $J_2, J_3, J_4, J_5$ & $T_2, T_4$ \\
\hline
$TS_2$ & $L_2$  &$J_1, J_2, J_3, J_5$ &  $T_2, T_3$ \\
\hline
$TS_3$ & $L_1, L_7$ & $J_2, J_4, J_5$ & $T_2 T_3$ \\
\hline
$TS_4$ & $L_2, L_3$ & $J_3, J_5, J_6$ & $T_1$  \\
\hline
$\ldots$ &  & &   \\
\hline
$TS_{76}$ & $L_1, L_3, L_5, L_8, L_9 $ & $J_2, J_4$ & $T_3$  \\
\hline

\end{tabular}
\label{tab10}
\end{center}
\end{table}
We now implement Algorithm 1 over the data to collect the agro-triples. It is observed that a total of seventy six agro-triples were generated. Some of them are shown in Table~\ref{tab10}.

Now we are ready to implement Algorithm 2 for the statio-temporal lattice creation.

The lattice structure so created with all the properties mentioned in the previous section is shown in Fig.~\ref{fig}. Finally Algorithm~3 is implemented and a total of twenty six associative rules were generated with a support of 0.7 and a confidence of 0.8. 

The asymptotic time complexity analysis of the algorithms are now discussed. Algorithm~1 bears a time complexity of $O(L \times J \times T)$. Algorithm~2 has a complexity of $O(L \times J \times T \times |S|)$, where $|S|$ is the size of the spatio-temporal lattice structure. Finally Algorithm~3 also take a time of $O(L \times J \times T)$.

\section{Conclusion}

With the integration of service-oriented architecture services, precision farming has become a huge opportunity in agriculture. It's imperative that new approaches for processing raw data and extracting valuable information be developed.
This is particularly true for the field of agriculture. Regardless of the fact that the data are constantly changing day by day, meaningful data extraction is used by many disciplines, including IoT-based enterprises, the automation sector, corporations, and many more. With the use of ontologies, it is possible to extract relevant information from a set of data by looking at their relationships. Ontology may be used for a variety of purposes, including decision support systems, expert systems, and the reuse of domain knowledge.

In this paper, we have suggested the use of a lattice structure for knowledge representation of the data collected from the IoT devices. The lattice structure is created from spatio-temporal data and finally rules were generated based on the properties of the lattice for reasoning. In future, we tend to collect real time data from sensors and implement the same over the algorithms described. 

\section*{Acknowledgement}
This research is partially supported by the NSF Grant 2025682 at Tennessee Tech University, USA.

\bibliographystyle{plain}

\end{document}